\pdfoutput=1

\documentclass[11pt]{article}

\usepackage{naacl2021}

\usepackage{times}
\usepackage{latexsym}

\usepackage[T1]{fontenc}

\usepackage[utf8]{inputenc}

\usepackage{microtype}

\usepackage{graphicx}
\usepackage{booktabs}
\usepackage{multirow}
\usepackage{multicol}

\title{Probing Word Translations in the Transformer and\\ Trading Decoder for Encoder Layers}

\author{Hongfei Xu$^{1,2}$\ \ \ \ Josef van Genabith$^{1,2}$\ \ \ \ Qiuhui Liu$^3$\thanks{\ \ \ \ Corresponding author.}\ \ \ \ Deyi Xiong$^{4}$\\
$^1$Saarland University / Saarland, Germany\\
$^2$German Research Center for Artificial Intelligence / Saarland, Germany\\
$^3$China Mobile Online Services / Henan, China\\
$^4$Tianjin University / Tianjin, China\\
hfxunlp@foxmail.com,
Josef.Van\_Genabith@dfki.de,\\
liuqhano@foxmail.com,
dyxiong@tju.edu.cn}

\begin{document}
\maketitle
\begin{abstract}
Due to its effectiveness and performance, the Transformer translation model has attracted wide attention, most recently in terms of probing-based approaches. Previous work focuses on using or probing source linguistic features in the encoder. To date, the way word translation evolves in Transformer layers has not yet been investigated. Naively, one might assume that encoder layers capture source information while decoder layers translate. In this work, we show that this is not quite the case: translation already happens progressively in encoder layers and even in the input embeddings. More surprisingly, we find that some of the lower decoder layers do not actually do that much decoding. We show all of this in terms of a probing approach where we project representations of the layer analyzed to the final trained and frozen classifier level of the Transformer decoder to measure word translation accuracy. Our findings motivate and explain a Transformer configuration change: if translation already happens in the encoder layers, perhaps we can increase the number of encoder layers, while decreasing the number of decoder layers, boosting decoding speed, without loss in translation quality? Our experiments show that this is indeed the case: we can increase speed by up to a factor $2.3$ with small gains in translation quality, while an $18$-$4$ deep encoder configuration boosts translation quality by $+1.42$ BLEU (En-De) at a speed-up of $1.4$.
\end{abstract}

\section{Introduction}

Neural Machine Translation (NMT) has achieved great success in the last few years. The popular Transformer \cite{vaswani2017attention} model, which outperforms previous RNN/CNN based translation models \cite{bahdanau2014neural,gehring2017convolutional}, is based on multi-layer self-attention networks and can be parallelized effectively.

\begin{figure*}[t]
	\centering
	\includegraphics[width=1.7\columnwidth]{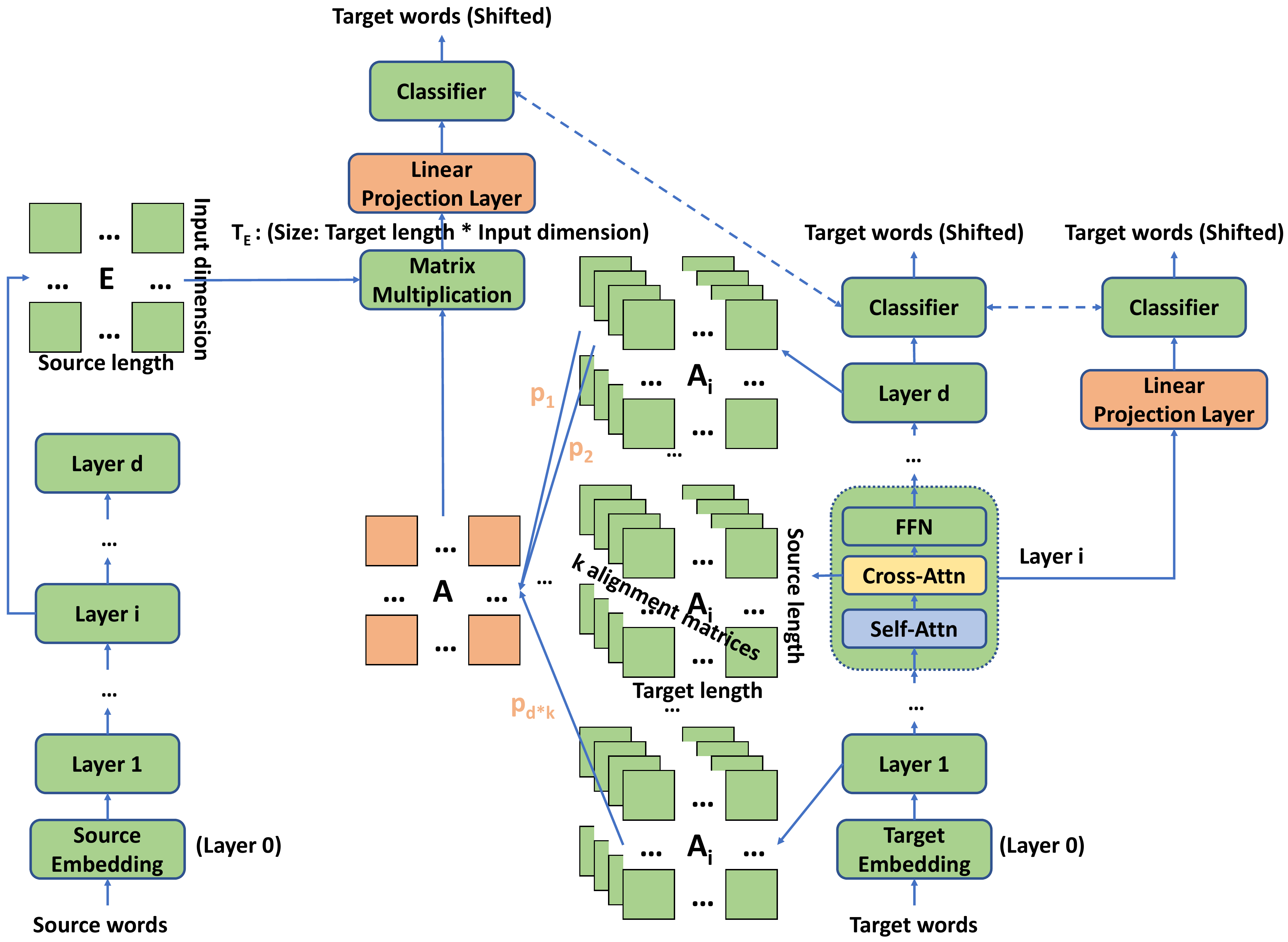}
	\caption{Analyzing word translations of Transformer layers. Green indicates layers of the trained Transformer model frozen for analysis. Orange indicates parameters of the linear projection layer and weights of alignment matrices $A_i$ trained on the training set. Dashed arrows indicate shared modules. When analyzing the separate effects of source contexts or decoding history in a decoder layer, one of the cross-attention (in yellow) or self-attention sub-layers (in blue) of the analyzed decoder layer are bypassed by a residual connection (Section \ref{sec:ana:dec}). Layers are independently analyzed. Target words (Shifted): the reference translation is one-position right-shifted compared to decoder input, i.e., predicting the next word with the current word as input.}
	\label{fig:ana}
\end{figure*}

Recently, a wide range of studies related to the Transformer have been conducted. For example, \newcite{bisazza2018lazy} perform a fine-grained analysis of how various source-side morphological features are captured at different levels of an NMT encoder. Surprisingly, they do not find any correlation between the accuracy of source morphology encoding and translation quality. Morphological features are only captured in context and only to the extent that they are directly transferable to target words. \newcite{voita2019bottom} study how information flows across Transformer layers and find that representations differ significantly depending on the objectives (machine translation, standard left-to-right language models and masked language modeling). \newcite{tang2019encoders} find that encoder hidden states outperform word embeddings significantly in word sense disambiguation. However, to the best of our knowledge, to date there is no study about how the Transformer translation model transforms individual source tokens into corresponding target tokens (i.e., word translations), and specifically, which role each Transformer layer plays in word translation, and at which layer a word is translated.

To investigate the roles of Transformer layers in translation, in this paper, we adopt probing approaches \cite{yossi2017fine,hupkes2018visualisation,conneau2018cram} and propose to measure the word translation accuracy of output representations of individual Transformer layers by probing how capable they are at translating words. Probing uses linear classifiers, referred to as ``probes'', where a probe can only use the hidden units of a given intermediate layer as discriminating features. Moreover, these probes cannot affect the training phase of a model, and they are generally added after training \cite{alain2017understanding}. In addition to analyzing the role of each encoder/decoder layer, we also analyze the contribution of the source context and the decoding history in translation by testing the effects of the masked self-attention sub-layer and the cross-attention sub-layer in decoder layers.

We present empirical results for how word translation is performed in each encoder/decoder layer, and how the alignment modeling (cross-attention sub-layers) and language modeling (masked self-attention sub-layers) contribute to the performance in each decoder layer. Our analysis demonstrates how word translation evolves across encoder/decoder layers and provides insights into the impact of the source ``encoding'' and the decoding history on the translation of target tokens. It reveals the existence of target translations in encoder states (and even source word embeddings) and the translation performed by encoder layers.

Based on our findings, we show that the proper use of more encoder layers with fewer decoder layers can significantly boost decoding speed without harming quality. Recently, \newcite{kasai2020deep} independently and similar to our encoder-decoder layer trading approach, compare the performance and speed of a 12-layer encoder 1-layer decoder with Non-Autoregressive Translation (NAT) approaches, and show that a one-layer autoregressive decoder can yield state-of-the-art accuracy with comparable latency to strong non-autoregressive models. Our analysis explains why using a deep encoder with a shallow decoder is feasible, and we show that some encoder-decoder depth configurations deliver both increased speed and increased translation quality.

\section{Probing Layer-wise Word Translation}

To analyze word translation accuracy of the Transformer, we first freeze a trained Transformer model so its behavior is consistent in how it performs in translation during our analysis.  We then extract output representations of the particular layer analyzed, apply a linear projection layer to extract features related to translation and feed the projected representations to the frozen decoder classifier of the trained Transformer. Our approach is minimally invasive in that only the linear projection layer and the weights of the alignment matrix $A$ responsible for combining frozen cross-attention alignment matrices from the decoder are trained and updated on the training set, with the original Transformer being frozen.  Thus the projection layer will only transform between vector spaces without generating new features for the word translation, and the alignment matrix $A$ will only combine frozen cross-attention alignment matrices. A high-level illustration of our analysis approach for encoder/decoder layers is shown in Figure \ref{fig:ana}.

\subsection{Analysis of Encoder Layers}

Analyzing word translation accuracy of encoder layers requires us to align source tokens with corresponding target tokens. We use the frozen alignment matrices computed by cross-attention sub-layers in decoder layers to align source tokens with target tokens (Figure \ref{fig:ana}). As there are multiple matrices produced by each sub-layer (due to the multi-head attention mechanism) and multiple decoder layers, we have to ensemble them into one matrix of high alignment accuracy using weights. Assume there are $d$ decoder layers with $k$ attention heads in each multi-head attention sub-layer, which results in $d * k$ alignment matrices $A_1, ..., A_{d * k}$. We use a $d * k$ dimension weight vector $w$ to combine all attention matrices. The weight vector is normalized by softmax to a probability distribution $p$:

\begin{equation}
     {p_i} = \frac{{e^{{w_i}}}}{{\sum\limits_{j = 1}^{d * k} {{e^{{w_j}}}} }}
\end{equation}

\noindent where $i$ indicates the $i$th element in $w$.

Then we use $p$ as the weights of the corresponding attention matrices and merge them into one alignment matrix $A$.

\begin{equation}
     A = \sum\limits_{i = 1}^{d*k} {{A_i}*{p_i}}
\end{equation}

$w$ is trained with the linear projection layer through backpropagation on the frozen Transformer.

After we obtain the alignment matrix $A$, instead of selecting the target token with the highest alignment weight as the translation of a source token, we perform matrix multiplication between the encoded source representations $E$ (size: source sentence length $*$ input dimension) and the alignment matrix $A$ (size: source sentence length $*$ target sentence length) to transform/re-order source representations to the target side $T_E$:

\begin{equation}
     {T_E} = {A^T} \times E
\end{equation}

\noindent where $A^T$ and $\times$ indicate the transpose of $A$ and matrix multiplication.

Thus $T_E$ has the same length as the gold translation sequence, and the ground-truth target sequence can be used directly as the translation represented by $T_E$.

Though source representations are transformed to the target side, we suggest this does not involve any target side information as the pre-trained Transformer is frozen and the transformation does not introduce any representation from the decoder side. We do not retrieve target tokens with the highest alignment score as word translations of corresponding source tokens because translation may involve zero/one/multiple source token(s) to zero/one/multiple target token(s) alignments, and we suggest that using a soft alignment (attention weights) may lead to more reliable gradients than a hard alignment.

\subsection{Analysis of Decoder Layers}
\label{sec:ana:dec}

The analysis of the prediction accuracy of the decoder is simpler than the encoder, as we can directly use the shifted target sequence (teacher forcing) without the requirement to bridge different sequence lengths between the source sentence and the target while analyzing the encoder. We use the output representations of the analyzed layer, and evaluate its prediction accuracy after projection.

However, as studied by \newcite{li2019word}, the decoder involves two kinds of ``translation''. One (performed by the self-attention sub-layer) translates the history token sequence to the next token, another (performed by the cross-attention sub-layer) translates by attending source tokens. We additionally analyze the effects of these two kinds of translation on predicting accuracy by dropping the corresponding sub-layer (either cross- or masked self-attention) of the analyzed decoder layer (i.e., we only compute the other sub-layer and the feed-forward layer where only the residual connection is kept as the computation of the skipped sub-layer).

\begin{table*}[t]
  \centering
    \begin{tabular}{r|rr|rrrrrr}
    \toprule
    \multicolumn{1}{c}{\multirow{3}[6]{*}{Layer}} & \multicolumn{2}{|c|}{Encoder} & \multicolumn{6}{c}{Decoder} \\
          & \multicolumn{1}{c}{\multirow{2}[4]{*}{Acc}} & \multicolumn{1}{c|}{\multirow{2}[4]{*}{$\Delta$}} & \multicolumn{1}{c}{\multirow{2}[4]{*}{Acc}} & \multicolumn{1}{c}{\multirow{2}[4]{*}{$\Delta$}} & \multicolumn{2}{c}{-Self attention} & \multicolumn{2}{c}{-Cross attention} \\
          &       &       &       &       & \multicolumn{1}{c}{Acc} & \multicolumn{1}{c}{$\Delta$} & \multicolumn{1}{c}{Acc} & \multicolumn{1}{c}{$\Delta$} \\
    \midrule
    0     & 40.73  &       & 13.72  & \multicolumn{5}{c}{} \\

    1     & 41.85  & 1.12  & 20.52  & 6.80  & 17.46  & -3.06  & 16.47  & -4.05  \\
    2     & 43.75  & 1.90  & 26.06  & 5.54  & 21.03  & -5.03  & 22.91  & -3.15  \\
    3     & 45.49  & 1.74  & 34.13  & 8.07  & 26.68  & -7.45  & 27.79  & -6.34  \\
    4     & 47.14  & 1.65  & 55.00  & 20.87  & 39.43  & -15.57  & 35.32  & -19.68  \\
    5     & 48.35  & 1.21  & 66.14  & 11.14  & 62.60  & -3.54  & 55.84  & -10.30  \\
    6     & 49.22  & 0.87  & 70.80  & 4.66  & 70.13  & -0.67  & 69.03  & -1.77  \\
    \bottomrule
    \end{tabular}%
  \caption{Word translation accuracy of Transformer layers on the WMT 14 En-De task.}
  \label{tab:wacce6d6}%
\end{table*}%

\begin{table*}[t]
  \centering
    \begin{tabular}{r|rr|rrrrrr}
    \toprule
    \multicolumn{1}{c}{\multirow{3}[6]{*}{Layer}} & \multicolumn{2}{|c|}{Encoder} & \multicolumn{6}{c}{Decoder} \\
          & \multicolumn{1}{c}{\multirow{2}[4]{*}{Acc}} & \multicolumn{1}{c|}{\multirow{2}[4]{*}{$\Delta$}} & \multicolumn{1}{c}{\multirow{2}[4]{*}{Acc}} & \multicolumn{1}{c}{\multirow{2}[4]{*}{$\Delta$}} & \multicolumn{2}{c}{-Self attention} & \multicolumn{2}{c}{-Cross attention} \\
          &       &       &       &       & \multicolumn{1}{c}{Acc} & \multicolumn{1}{c}{$\Delta$} & \multicolumn{1}{c}{Acc} & \multicolumn{1}{c}{$\Delta$} \\
    \midrule
    0    & 41.87 &       & 16.26 & \multicolumn{5}{c}{} \\
    1    & 43.61 & 1.74 & 25.73 & 9.47 & 23.31 & -2.42 & 18.89 & -6.84 \\
    2    & 45.26 & 1.65 & 32.55 & 6.82 & 27.10 & -5.45 & 26.82 & -5.73 \\
    3    & 46.68 & 1.42 & 40.80 & 8.25 & 34.05 & -6.75 & 32.84 & -7.96 \\
    4    & 47.88 & 1.20 & 55.60 & 14.80 & 47.29 & -8.31 & 40.48 & -15.12 \\
    5    & 48.73 & 0.85 & 64.39 & 8.79 & 62.41 & -1.98 & 55.69 & -8.70 \\
    6    & 49.39 & 0.66 & 67.10 & 2.71 & 66.93 & -0.17 & 66.31 & -0.79 \\
    \bottomrule
    \end{tabular}%
  \caption{Word translation accuracy of Transformer layers on the WMT 15 Cs-En task.}
  \label{tab:cswacce6d6}%
\end{table*}%

\section{Analysis Experiments}

\subsection{Settings}

We first trained a Transformer base model for our analysis on the popular WMT 14 English to German news translation task to compare with \newcite{vaswani2017attention}. We employed a $512 * 512$ parameter matrix as the linear projection layer. The source embedding matrix, the target embedding matrix and the weight matrix of the classifier were tied. Parameters were initialized under the Lipschitz constraint \cite{xu2020lipschitz} to ensure the convergence of deep encoders. We implemented our approaches based on the Neutron implementation \citep{xu2019neutron} of the Transformer translation model.

We applied joint Byte-Pair Encoding (BPE) \cite{sennrich2016neural} with $32k$ merge operations. We only kept sentences with a maximum of $256$ sub-word tokens for training. The concatenation of newstest 2012 and newstest 2013 was used for validation and newstest 2014 as the test set.

The number of warm-up steps was set to $8k$.\footnote{\url{https://github.com/tensorflow/tensor2tensor/blob/v1.15.4/tensor2tensor/models/transformer.py\#L1818}.} The model was trained for $100k$ training steps with around $25k$ target tokens in each batch. We followed all the other settings of \newcite{vaswani2017attention}.

We averaged the last $5$ checkpoints saved with an interval of $1,500$ training steps. For decoding, we used a beam size of $4$, and evaluated tokenized case-sensitive BLEU.\footnote{\url{https://github.com/moses-smt/mosesdecoder/blob/master/scripts/generic/multi-bleu.perl}.} The averaged model achieved a BLEU score of $27.96$ on the test set.

The projection matrix and the weight vector $w$ of $48$ elements for alignment were trained on the training set with the frozen Transformer. We monitored the accuracy on the development set, and report results on the test set.

\subsection{Analysis}
\label{subsec:ana}

The analysis results of the trained Transformer are shown in Table \ref{tab:wacce6d6}. Layer $0$ stands for the embedding layer. ``Acc'' indicates the prediction accuracy. ``-Self attention'' and ``-Cross attention'' in the decoder layer analysis mean bypassing the computation of the masked self-attention sub-layer and the cross-attention sub-layer respectively of the analyzed decoder layer using a residual connection. In our layer analysis of the encoder and decoder, ``$\Delta$'' indicates improvements in word translation accuracy of the analyzed layer over the previous layer. While analyzing the self-attention and cross-attention sub-layers, ``$\Delta$'' is the accuracy loss when we remove the computation of the corresponding sub-layer.

The results of the encoder layers in Table \ref{tab:wacce6d6} show that: 1) encoder layers already perform word translation, and the translation even starts at the embedding layer with unexpectedly high accuracy. 2) With the stacking of encoder layers, the word translation accuracy improves, and improvements brought about by different layers are relatively similar, indicating that all encoder layers are useful.

Surprisingly, analyzing decoder layers, Table \ref{tab:wacce6d6} shows that: 1) shallow decoder layers (0, 1, 2 and 3) perform significantly worse compared to the corresponding encoder layers (all the way up until the $4$th decoder layer, where a word translation accuracy which surpasses the embedding layer of the encoder is achieved); 2) The improvements brought about by different decoder layers are quite different. Specifically, the relative performance increases between the low-performance decoder layers (0, 1, 2 and 3) are low as well, while layers 4 and 5 bring more improvements than the others.

While analyzing the effects of the source context (``-Cross attention'' prevents informing translation by the source ``encoding'') and the decoding history (the self-attention sub-layer is responsible for the target language re-ordering, and ``-Self attention'' prevents using the decoding history in the analyzed decoder layer), Table \ref{tab:wacce6d6} shows that in shallow decoder layers (layer $1$-$3$), the decoding history is as important as the source ``encoding'', while in deep decoder layers, the source ``encoding'' plays a more vital role than the decoding history. Overall, our results provide new insights on the importance of translation already performed by the encoder.

Since the English-German translation shares many sub-words naturally ($\sim$$13.89\%$ source subwords including punctuations exist in the subword set of the corresponding target translation in the training set), we additionally provide results on the WMT 15 Cs-En task in Table \ref{tab:cswacce6d6}. Table \ref{tab:cswacce6d6} confirms our observations reported in Table \ref{tab:wacce6d6}.

\newcite{zhang2018language,hewitt2019designing,voita2020information} articulate concerns about analyses with probing accuracies, as differences in accuracies fail to reflect differences in representations in several ``sanity checks''. Specifically, \newcite{zhang2018language} compare probing scores for trained models and randomly initialized ones, and observe reasonable differences in the scores only when reducing the amount of classifier training data. However, we argue that in our work, we use the frozen classifier of the pre-trained Transformer decoder as our probing classifier, and the introduced linear projection, as well as the alignment matrix $A$, are much smaller and weaker than the frozen classifier and the rest of the frozen Transformer components. Thus we suggest that our approach is minimally invasive and that our analysis is less likely to be seriously affected by this issue even though we use a large training set. To empirically verify this, we apply our analysis approach on a randomly initialized encoder and evaluate word translation accuracies obtained by the source embedding layer and last encoder layer, while the alignment between the source and the target is still from the pre-trained model. Both the source embedding layer and the last encoder layer resulted in the same accuracy of $23.66$. Compared to the corresponding values ($40.73$ and $49.22$) in Table \ref{tab:wacce6d6}, the gap between the randomly initialized layers and the pre-trained layers in accuracy is significant, and the gap between accuracy improvements from the representation extracted from the source embedding layer and propagated through all intermediate layers to the last encoder layer of pre-trained layers ($8.49$) and randomly initialized layers ($0.00$) is also significant. Thus, we suggest our analysis is robust.

\begin{table}[t]
  \centering
    \begin{tabular}{lrrrr}
    \toprule
    Layer & \multicolumn{1}{c}{BLEU 1} & \multicolumn{1}{c}{$\Delta$} & \multicolumn{1}{c}{BLEU} & \multicolumn{1}{c}{$\Delta$} \\
    \midrule
    \multicolumn{1}{r}{0} & 33.1  &       & 7.92  &  \\
    \multicolumn{1}{r}{1} & 35.7  & 2.6   & 8.99  & 1.07 \\
    \multicolumn{1}{r}{2} & 41.0  & 5.3   & 11.05 & 2.06 \\
    \multicolumn{1}{r}{3} & 43.3  & 2.3   & 11.89 & 0.84 \\
    \multicolumn{1}{r}{4} & 46.8  & 3.5   & 13.13 & 1.24 \\
    \multicolumn{1}{r}{5} & 48.1  & 1.3   & 13.34 & 0.21 \\
    \multicolumn{1}{r}{6} & 48.6  & 0.5   & 13.45 & 0.11 \\
    FULL  & 62.0  & 13.4  & 33.26 & 19.81 \\
    \bottomrule
    \end{tabular}%
  \caption{Translation performance of encoder layers on the WMT 14 En-De task.}
  \label{tab:enctbleu}%
\end{table}%

\subsection{Translation from Encoder Layers without Using Decoder Layers}
\label{subsec:tenc}

Since our approach extracts features for translation from encoder states while analyzing them, is it possible to perform word translation with only these features from encoder layers without using the decoder except the frozen classifier?

To test this question, we feed output representations from an encoder layer to the corresponding linear projection layer, and feed the output of the linear projection layer directly to the frozen decoder classifier, and retrieve tokens with the highest probabilities as ``translations''. Even though such ``translations'' from encoder layers have the same length and the same word order as source sentences, individual source tokens are translated to the target language to some extent. We evaluated BPEized \footnote{Since there is no re-ordering of the target language performed, which makes the merging of translated sub-word units in the source sentence order pointless.} case-insensitive BLEU and BLEU 1 (1-gram BLEU, indicates the word translation quality), and results are shown in Table \ref{tab:enctbleu}. ``FULL'' is the performance of the whole Transformer model (decoding with a beam size of 4). ``$\Delta$'' means the improvements obtained by the introduced layer (or the decoder for ``FULL'') over the previous layer.

Table \ref{tab:enctbleu} shows that while there is a significant gap in BLEU scores between encoder layers and the full Transformer, the gap in BLEU 1 is relatively smaller than in BLEU. It is reasonable that encoder layers achieve a comparably high BLEU 1 score but a low BLEU score overall, as they perform word translation in the same order as the source sentence without any word re-ordering of the target language. We suggest that the BLEU 1 score achieved by only the source embedding layer (i.e., translating with only embeddings) is surprising and worth noting.

\subsection{Discussion}

Our probing approach involves crucial information from the decoder (encoder-decoder attention from all decoder layers). However, we argue that probe training requires supervision. For the decoder, we can directly use gold references. On the encoder side, parallel data does not provide word translations for source tokens, and we have to generate this data by aligning target tokens to source tokens. One choice is extracting alignments by taking an argmax of alignment matrices or using toolkits like fastalign \citep{dyer2013simple}. In this case, probe training does not involve attention matrices, but this has drawbacks: multiple/no target tokens may align to one source token. We use soft aggregation to preserve more information (other attention possibilities besides the highest are kept) and to alleviate error propagation. We argue that the use of attention matrices is only to bring supervision (word translations) from the target side to the source side, which is inevitable. Decoder representations cannot flow back to the frozen encoder.

Our paper also empirically reveals the impact of attention matrices: 1) In Section \ref{subsec:tenc}, where after the training of source probes, we decode target tokens with only encoder layers, the trained probe (without involving cross-attention networks) and the pre-trained classifier. 2) In the last paragraph of Section \ref{subsec:ana}, we train probes with alignment matrices from the pre-trained model but a frozen random encoder, showing the effects of cross-attention matrices on the probe.

\begin{table*}[t]
  \centering
    \begin{tabular}{crrrrrrr}
    \toprule
    \multicolumn{1}{c}{Model} & \multicolumn{2}{c}{Depth} & \multicolumn{1}{c}{\multirow{2}[4]{*}{BLEU}} & \multicolumn{1}{c}{\multirow{2}[4]{*}{Para. (M)}} & \multicolumn{3}{c}{Time} \\
     & \multicolumn{1}{c}{Encoder} & \multicolumn{1}{c}{Decoder} &       &       & \multicolumn{1}{c}{Train} & \multicolumn{1}{c}{Decode (/s)} & \multicolumn{1}{c}{Speed up} \\
    \midrule
	\multicolumn{1}{c}{\newcite{zhang2018accelerating}} & 6     & 6     & 28.13 & 74.97 & 40h09m & 29 & 1.52  \\
	\midrule
    \multirow{7}[2]{*}{Transformer} & 6     & 6     & 27.96 & 62.37 & 33h33m & 44 & 1.00  \\
     & 7     & 5     & 28.07 & 61.32 & 32h17m & 38 & 1.16  \\
     & 8     & 4     & 28.61 & 60.27 & 31h26m & 31 & 1.42  \\
     & 9     & 3     & 28.53 & 59.22 & 30h29m & 25 & 1.76  \\
     & 10    & 2     & 28.47 & 58.17 & 30h11m & 19 & 2.32  \\
     & 11    & 1     & 27.02 & 57.12 & 29h27m & 13 & 3.38  \\
	\cmidrule{2-8}
	 & 18    & 4     & 29.38 & 91.77 & 52h56m & 32 & 1.38  \\
    \bottomrule
    \end{tabular}%
  \caption{Effects of encoder/decoder depth on the WMT 14 En-De task. The decoding time is for the test set of $3,003$ sentences with a beam size of $4$.}
  \label{tab:edepth}%
\end{table*}%

\section{Trading Decoder for Encoder Layers}

\subsection{Motivation}

From our analysis of the 6-layer Transformer base model (Table \ref{tab:wacce6d6}), we find that in contrast to the improvements of the word translation accuracy with increasing depth on the encoder side, some decoder layers contribute significantly fewer improvements than others (i.e., layers 4 and 5 bring more word translation accuracy improvements than those from layers 1, 2, 3 and 6 in Table \ref{tab:wacce6d6}). This suggests that there might be more ``lazy'' layers in the decoder than in the encoder, which means that it might be easier to compress the decoder than the encoder, and further we conjecture that simply removing some decoder layers while adding the same number of encoder layers may even improve the translation quality of the transformer. Motivations targeting efficiency include:

\begin{itemize}
    \item Each decoder layer has one more cross-attention sub-layer than an encoder layer, and increasing encoder layers while decreasing the same number of decoder layers will reduce the number of parameters and computational cost;
    \item During inference, the decoder has to autoregressively compute the forward pass for every decoding step (the decoding of each target token), which prevents efficient parallelization, while encoder layers are non-autoregressively propagated and highly parallelized, and the acceleration caused by using fewer decoder layers with more encoder layers will be more significant in decoding, which is of practical value.
\end{itemize}

\subsection{Results and Analysis}

\begin{table*}[t]
  \centering
    \begin{tabular}{r|rr|cccccc}
    \toprule
    \multicolumn{1}{c}{\multirow{3}[6]{*}{Layer}} & \multicolumn{2}{|c|}{Encoder} & \multicolumn{6}{c}{Decoder} \\
          & \multicolumn{1}{c}{\multirow{2}[4]{*}{Acc}} & \multicolumn{1}{c|}{\multirow{2}[4]{*}{$\Delta$}} & \multirow{2}[4]{*}{Acc} & \multirow{2}[4]{*}{$\Delta$} & \multicolumn{2}{c}{-Self attention} & \multicolumn{2}{c}{-Cross attention} \\
          &       &       &       &       & \multicolumn{1}{c}{Acc} & \multicolumn{1}{c}{$\Delta$} & \multicolumn{1}{c}{Acc} & \multicolumn{1}{c}{$\Delta$} \\
    \midrule
    0     & 40.48 &       & \multicolumn{1}{r}{14.04} & \multicolumn{5}{c}{} \\
    1     & 41.29 & 0.81  & \multicolumn{1}{r}{37.42} & \multicolumn{1}{r}{23.38} & \multicolumn{1}{r}{25.56} & \multicolumn{1}{r}{-11.86} & \multicolumn{1}{r}{20.40} & \multicolumn{1}{r}{-17.02} \\
    2     & 43.00 & 1.71  & \multicolumn{1}{r}{68.77} & \multicolumn{1}{r}{31.35} & \multicolumn{1}{r}{62.01} & \multicolumn{1}{r}{-6.76} & \multicolumn{1}{r}{40.67} & \multicolumn{1}{r}{-28.10} \\
    3     & 44.07 & 1.07  & \multicolumn{6}{c}{\multirow{8}[16]{*}{}} \\
    4     & 45.86 & 1.79  & \multicolumn{6}{c}{} \\
    5     & 46.54 & 0.68  & \multicolumn{6}{c}{} \\
    6     & 47.46 & 0.92  & \multicolumn{6}{c}{} \\
    7     & 48.92 & 1.46  & \multicolumn{6}{c}{} \\
    8     & 49.58 & 0.66  & \multicolumn{6}{c}{} \\
    9     & 50.24 & 0.66  & \multicolumn{6}{c}{} \\
    10    & 50.35 & 0.11  & \multicolumn{6}{c}{} \\
    \bottomrule
    \end{tabular}%
  \caption{Word accuracy analysis on Transformer with $10$ encoder and $2$ decoder layers on the WMT 14 En-De task.}
  \label{tab:wacce10d2}%
\end{table*}%

We examine the effects of reducing the number of decoder layers while adding corresponding numbers of encoder layers, and results are shown in Table \ref{tab:edepth}. ``Speed up'' stands for the decoding acceleration compared to the 6-layer Transformer.

Table \ref{tab:edepth} shows that while the acceleration of trading decoder layers for encoder layers in training is small, in decoding it is significant. Specifically, the Transformer with $10$ encoder layers and $2$ decoder layers is $2.32$ times as fast as the 6-layer Transformer while achieving a slightly higher BLEU.

Can we use more than $12$ encoder layers with a shallow decoder to benefit both translation quality and inference speed? Table \ref{tab:edepth} shows that the $18$-$4$ model \footnote{A full grid search over configurations is tedious and expensive. We take inspiration from Table \ref{tab:edepth} where going from $5$ to $4$ decoder layers brings about the biggest relative jump in translation quality. We explored a few configurations and find that using more than $18$ encoder layers can still bring improvements, but the gains are relatively small.} brings about $+1.42$ BLEU improvements over the strong baseline, while being $1.38$ times as fast in decoding. Comparing the $18$-$4$ model to the $8$-$4$ model, the time cost for using $10$ more encoder layers only increases $1$ second for translating the test set, suggesting that autoregressive decoding speed is quite insensitive to the encoder depth.

Our results show that using more encoder layers with fewer but sufficient decoder layers can significantly boost the decoding speed with small gains in translation quality, and that a good choice in the distribution of encoder and decoder layers ($18$-$4$) can result in slightly faster decoding and a substantial increase in translation quality, which is simple but effective and valuable for back-translation \cite{sennrich2016improving} and production applications.

We present the word accuracy analysis results of the $10$ encoder layer - $2$ decoder layer Transformer on the En-De task in Table \ref{tab:wacce10d2}. Comparing Table \ref{tab:wacce10d2} with Table \ref{tab:wacce6d6}, we find that: 1) The differences in improvements ($1.71$ vs. $0.11$) brought by individual layers of the 10-layer encoder are larger than those of the 6-layer encoder ($1.90$ vs. $0.87$), indicating that there might now be some ``lazy'' layers in the 10-layer encoder; 2) Decreasing the depth of the decoder removes ``lazy'' decoder layers in the 6-layer decoder and makes decoder layers rely more on the source ``encoding'' (by comparing the effects of skipping the self-attention sub-layer and cross-attention sub-layer on performance).

\begin{table}[t]
  \centering
    \begin{tabular}{cclll}
    \toprule
    \multicolumn{2}{c}{Depth} & \multicolumn{1}{c}{\multirow{2}[0]{*}{En-De}} & \multicolumn{1}{c}{\multirow{2}[0]{*}{En-Fr}} & \multicolumn{1}{c}{\multirow{2}[0]{*}{Cs-En}} \\
    Encoder & Decoder &       &       &  \\
    \midrule
    \multicolumn{2}{c}{6} & 27.96 & 40.13 & 28.69 \\
    10    & 2     & 28.47 & 40.49 & 28.87 \\
	18    & 4     & \textbf{29.38$^\dag$} & \textbf{40.90$^\dag$} & \textbf{29.75$^\dag$} \\
    \bottomrule
    \end{tabular}%
  \caption{Verification of deep encoder and shallow decoder on WMT En-De, En-Fr and Cs-En tasks. $\dag$ indicates significance at $p<0.01$.}
  \label{tab:bleuprjl}%
\end{table}%

\subsection{Verification of Deep Encoder and Shallow Decoder on other Language Pairs}

To investigate how a deep encoder with a shallow decoder will perform in other tasks, we conducted experiments on the WMT 14 English-French and WMT 15 Czech-English news translation tasks in addition to the WMT 14 English-German task. Results on newstest 2014 (En-De/Fr) and 2015 (Cs-En) respectively are shown in Table \ref{tab:bleuprjl}.

Table \ref{tab:bleuprjl} shows that the $10$-$2$ model consistently achieves higher BLEU scores than the $6$-layer model, and the $18$-$4$ model consistently leads to significant improvements in all $3$ tasks.

\section{Related Work}

\paragraph{Analysis of NMT Models.} \newcite{belinkov2020linguistic} analyze the representations learned by NMT models at various levels of granularity and evaluate their quality through relevant extrinsic properties. \newcite{li2019word} analyze the word alignment quality in NMT and the effect of alignment errors on translation errors. They demonstrate that NMT captures word alignment much better for those words mostly contributed from the source than those from the target. \newcite{voita2019analyzing} evaluate the contribution of individual attention heads to the overall performance of the model and analyze the roles played by them in the encoder. \newcite{yang2019assessing} propose a word reordering detection task to quantify how well the word order information is learned by Self-Attention Networks and RNN, and reveal that although recurrence structure makes the model more universally effective on learning word order, learning objectives matter more in the downstream tasks such as machine translation. \newcite{tsai2019transformer} regard attention as applying a kernel smoother over the inputs with the kernel scores being the similarities between inputs, and analyze individual components of the Transformer's attention with the new formulation via the lens of the kernel. \newcite{tang2019encoders} find that encoder hidden states outperform word embeddings significantly in word sense disambiguation. \newcite{he2019towards} measure the word importance by attributing the NMT output to every input word and reveal that words of certain syntactic categories have higher importance while the categories vary across language pairs. \newcite{voita2019bottom} use canonical correlation analysis and mutual information estimators to study how information flows across Transformer layers. Early work by \newcite{bisazza2018lazy} performs a fine-grained analysis of how various source-side morphological features are captured at different levels of the NMT encoder. While they are unable to find any correlation between the accuracy of source morphology encoding and translation quality, they discover that morphological features are only captured in context and only to the extent that they are directly transferable to the target words, and suggest encoder layers are ``lazy''. Our analysis offers an explanation for their results as the translation already starts at the source embedding layer, and possibly source embeddings already represent linguistic features of their translations.

\paragraph{Analysis of BERT.} BERT \cite{devlin2019bert} uses the Transformer encoder, and analysis of BERT may provide valuable references for analyzing the Transformer. \newcite{jawahar2019bert} provide support that BERT networks capture structural information, and perform a series of experiments to unpack the elements of English language structure learned by BERT. \newcite{tenney2019bert} employ the edge probing task suite, and find that BERT represents the steps of the traditional NLP pipeline in an interpretable and localizable way, and that the regions responsible for each step appear in the expected sequence: POS tagging, parsing, NER, semantic roles, then coreference. \newcite{pires2019multilingual} present a large number of probing experiments, and show that Multilingual-BERT’s robust ability to generalize cross-lingually is underpinned by a multilingual representation.

\paragraph{Accelerating Decoding.} \newcite{zhang2018accelerating} propose average attention as an alternative to the self-attention network in the Transformer decoder to accelerate decoding. \newcite{wu2018pay} introduce lightweight convolution and dynamic convolutions. The number of operations required by their approach scales linearly in the input length, whereas self-attention is quadratic. \newcite{zhang2018speeding} apply cube pruning to neural machine translation to speed up translation. \newcite{zhang2018exploring} propose to adopt an n-gram suffix-based equivalence function into beam search decoding, which obtains similar translation quality with a smaller beam size, making NMT decoding more efficient. Non-Autoregressive Translation (NAT) \cite{gu2018nonautoregressive,libovicky2018end,wei2019imitation,shao2019retrieving,li2019hint,wang2019non,guo2019non} enables parallelized decoding, while there is still a significant quality drop compared to traditional autoregressive beam search, our findings on using more encoder layers might also be adapted to NAT. Recently, and independently of our work, \newcite{kasai2020deep} compare the performance and speed between a 12-layer encoder 1-layer decoder case with NAT approaches, and show that a one-layer autoregressive decoder yields state-of-the-art accuracy with comparable latency to strong non-autoregressive models. Our work explains why using a deep encoder with a shallow decoder is feasible, and we show that substantial increases in decoding speed are possible with small gains in translation quality, and that for some configurations (e.g., $18$-$4$) significant translation quality increases with modest increases in decoding speed are possible.

\section{Conclusion}

We propose approaches for the analysis of word translation accuracy of Transformer layers to investigate how translation is performed. To measure word translation accuracy, our approach trains a linear projection layer that bridges representations from the frozen pre-trained analyzed layer and the frozen pre-trained classifier. While analyzing encoder layers, our approach additionally learns a weight vector to merge multiple attention matrices into one, and transforms the source ``encoding'' to the target shape by multiplying the merged alignment matrix. Both the linear projection layer and the weight vector are trained on the frozen Transformer. This is minimally invasive, and training the new parameters does not account for the findings reported. For the analysis of decoder layers, we additionally analyze the effects of the source context and the decoding history in word prediction through bypassing the corresponding cross- and self-attention sub-layers. Our findings motivate and explain the benefits of trading decoder for encoder layers in our approach and that of \newcite{kasai2020deep}.

Our analysis is the first to reveal the existence of target translations performed by encoder layers (including the source embedding layer). We show that increasing encoder depth while removing decoder layers can lead to significant BLEU improvements while boosting the decoding speed.

\section*{Acknowledgements}

We thank anonymous reviewers for their insightful comments. Hongfei Xu acknowledges the support of China Scholarship Council ([2018]3101, 201807040056). Josef van Genabith is supported by the German Federal Ministry of Education and Research (BMBF) under funding code 01IW20010 (CORA4NLP). Deyi Xiong is supported by the National Natural Science Foundation of China (Grant No. 61861130364), the Natural Science Foundation of Tianjin (Grant No. 19JCZDJC31400) and the Royal Society (London) (NAF$\backslash$R1$\backslash$180122).

\bibliography{custom}
\bibliographystyle{acl_natbib}

\end{document}